# IEEE Copyright Notice

# Knowledge Models for Cancer Clinical Practice Guidelines : Construction, Management and Usage in Question Answering

Pralaypati Ta*†, Bhumika Gupta*†, Arihant Jain*†, Sneha Sree C*†, Keerthi Ram*, Mohanasankar Sivaprakasam*†
*Healthcare Technology Innovation Centre (HTIC), Indian Institute of Technology Madras, India
†Department of Electrical Engineering, Indian Institute of Technology Madras, India
Email: pralaypati@htic.iitm.ac.in

*Abstract*— An automated knowledge modeling algorithm for Cancer Clinical Practice Guidelines (CPGs) extracts the knowledge contained in the CPG documents and transforms it into a programmatically interactable, easy-to-update structured model with minimal human intervention. The existing automated algorithms have minimal scope and cannot handle the varying complexity of the knowledge content in the CPGs for different cancer types. This work proposes an improved automated knowledge modeling algorithm to create knowledge models from the National Comprehensive Cancer Network (NCCN) CPGs in Oncology for different cancer types. The proposed algorithm has been evaluated with NCCN CPGs for four different cancer types. We also proposed an algorithm to compare the knowledge models for different versions of a guideline to discover the specific changes introduced in the treatment protocol of a new version. We created a question-answering (Q&A) framework with the guideline knowledge models as the augmented knowledge base to study our ability to query the knowledge models. We compiled a set of 32 question-answer pairs derived from two reliable data sources for the treatment of Non-Small Cell Lung Cancer (NSCLC) to evaluate the Q&A framework. The framework was evaluated against the question-answer pairs from one data source, and it can generate the answers with 54.5% accuracy from the treatment algorithm and 81.8% accuracy from the discussion part of the NCCN NSCLC guideline knowledge model.

## I. Introduction

The clinical practice guidelines (CPGs) for cancer diseases provide care pathway guidelines for treating cancer patients. The National Comprehensive Cancer Network's (NCCN) Clinical Practice Guidelines in Oncology [1] are widely used by oncologists to manage cancer patients. The knowledge contained in the NCCN CPGs is frequently updated due to ongoing research in cancer therapy. It is desirable to have a machine-readable knowledge model to manage this knowledge [2] so that the model can be updated quickly and queried accurately. To capture the knowledge contained in the CPGs, several computer-interpretable guidelines (CIGs) models have been proposed in the literature [3], [4]. However, the suggested CIG models have not been widely used in reality; this could be because of the difficulty in transforming CPGs into CIGs and the scarcity of related tools [2], [5]. In our previous work [6], we proposed an automated process for extracting the knowledge components from the NCCN CPGs and creating a knowledge model using the open standard JSON-LD format. The method was validated using the NCCN treatment guidelines for Non-Small Cell Lung Cancer (NSCLC). NCCN publishes treatment guidelines for more than 60 different cancer types. The complexity of the knowledge content of the guideline documents varies depending on the type of cancer. The automated algorithm to create knowledge models should be able to faithfully extract the knowledge components from the NCCN CPG documents for all possible cancer types, irrespective of the complexity of the document content.

The NCCN oncology guidelines are updated regularly as information regarding cancer treatment advances. As the complexity of the NCCN guidelines increases significantly over time [7], it's necessary to have an automated method to identify the updates in a new version. A page-by-page overview of the modifications incorporated is included in a new version of an NCCN guideline document. However, it would be more convenient if the precise location of the modification on the page could be detected. A before-and-after comparison is also quite helpful in understanding the differences.

One of the primary purposes of knowledge modeling CPGs is to query the guideline knowledge programmatically to find the relevant information instead of manually finding it. It's desirable to have a question-answering (QA) system that can answer natural language questions from the guideline knowledge models. With the recent advances in the pre-trained large language models (LLM), the performance of the QA systems, including the medical QA systems, has improved significantly [8] [9]. As argued in [10] and [11], an evidence-based QA framework enhances the trustworthiness of a medical QA system. For the rapidly changing guideline knowledge, a retrieval augmented generation (RAG) [12] based LLM application can produce evidence-based responses using the guideline knowledge models as the augmented knowledge base.

The following are the primary contributions of this work:

- We created an improved version of the automated knowledge extraction technique proposed before [6], which can extract knowledge components from the NCCN guidelines for various cancer types. In addition to NSCLC, the algorithm has been evaluated with NCCN guidelines for three other common cancer types: Breast Cancer, Ovarian Cancer, and Prostate Cancer.

- We proposed an algorithm to compare the knowledge models of two different versions of an NCCN guideline. The algorithm can generate a page-wise detailed comparison result mentioning the exact location of the change in the page along with the type of the changes: addition, deletion, or modification. It assigns a score to the modified content to quickly identify how much it has been modified. The result also shows a before-and-after view of the changed contents.

- We developed a question-answering system, a Retrieval Augmented Generation (RAG) based LLM application, to answer natural language questions from the guideline knowledge model. We fine-tuned an embedding model to improve its domain-specific retrieval capability in the RAG pipeline by generating synthetic question-answers from the guideline content. To evaluate the QA application, we constructed a dataset of 32 question-answer pairs for the treatment of Non-Small Cell Lung Cancer (NSCLC) from two sources: 1) the *patient resources offered by the American Society of Clinical Oncology (ASCO)* [13] 2) the *patient resources offered by the National Cancer Institute (NCI)* [22]

## II. METHODOLOGY

### A. Knowledge Extraction for Different Cancer Types

NCCN guideline documents contain an algorithm section that depicts the treatment pathways graphically. The text blocks of treatment paths, called nodes, are joined by directed edges to show the sequential ordering. For different cancer types, the guideline documents use various forms of directed edges to illustrate the complex treatment algorithm, e.g., fan-out edges, multi-segment directed edges, and the edges connected to one after another. A faithful extraction of the edges is required to establish the relation between the nodes correctly. Fig. 1 shows steps to process the lines and directed edges.

The discussion section of the NCCN guidelines includes a detailed explanation of the recommendations given in the algorithm section, supported by clinical data and scientific reasoning. It primarily consists of textual information structured into sections, sub-sections, and paragraphs. The textual data was extracted paragraph by paragraph by *recognizing the structure of the guideline documents* using heuristics-based algorithms. Fig. 2 shows the steps in extracting the paragraphs from the discussion section.

### B. Knowledge Model Comparison

The knowledge model for the algorithm section of the NCCN guideline contains various types of nodes and the relations between them. A page-by-page comparison of the node contents is carried out to compare two different versions of such models. The hash of the node contents is matched to find the unchanged nodes. For nodes with varying hash values, a pair-wise *edit distance-based similarity score* between the node contents is calculated. If the score is above an upper threshold, the nodes are considered similar; if the score is below a lower threshold, categorize the node as an added or removed node; otherwise, consider it a modified node.

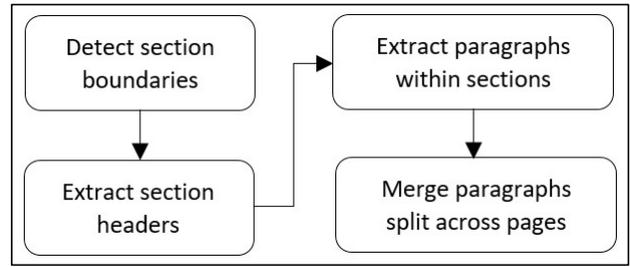

Fig. 2. Text extraction from discussion section.

### C. Question Answering Framework

The question-answering framework utilizes the algorithm section and the discussion section content as two distinct augmented knowledge bases. As a result, two RAG-based question-answering frameworks were created. Fig. 3 shows the components of the question-answering framework using the knowledge model graph paths from the algorithm section of the guideline. Initially, 150 treatment nodes were identified programmatically by comparing the labels, texts and other enrichment information [6] available in the knowledge model of NCCN guidelines for NSCLC version 3.2022. We manually reviewed the detected treatment nodes and removed those incorrectly recognized, yielding **121 treatment nodes**. Next, the knowledge model graph's *shortest paths* between the treatment nodes and the start node were determined. All the nodes in a path specify the different patient conditions and evaluation results that lead to the specific treatment node. The content of all such nodes in a path, except the last treatment node, is stored in a document. The most relevant documents are retrieved using a *hybrid retrieval* mechanism [19] for a given query. A *dense* retriever obtains semantically matching documents, while a *sparse* retriever is used to get lexically related documents. The retrieved documents are combined using the *reciprocal rank fusion (RRF)* algorithm, and the top-scoring two documents are obtained from them. The corresponding treatment nodes are considered answers to the question.

Fig. 4 shows the question-answering pipeline using the content from the discussion section. As in the QA framework for the algorithm section, both a dense and sparse index were created using the paragraphs of the discussion section. The paragraphs longer than the semantic embedding model's maximum input sequence length were divided into parts and an embedding vector was computed for each part. *The embedding for the whole paragraph was calculated by averaging the embedding of the smaller parts*. After combining using RRF, two top-scoring paragraphs were passed along with the question to an LLM. The LLM generates the final answer conditioned on the input paragraphs.

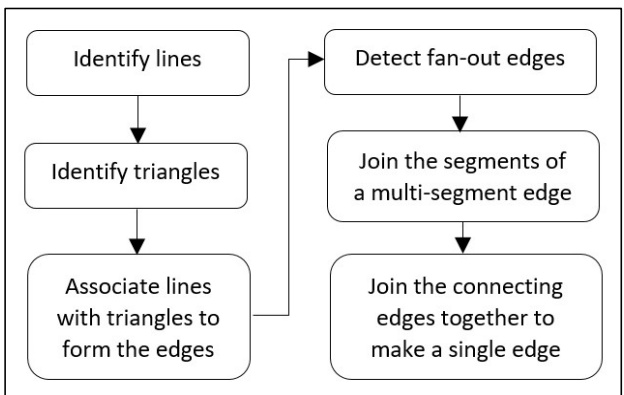

Fig. 1. Steps to process different form of edges.

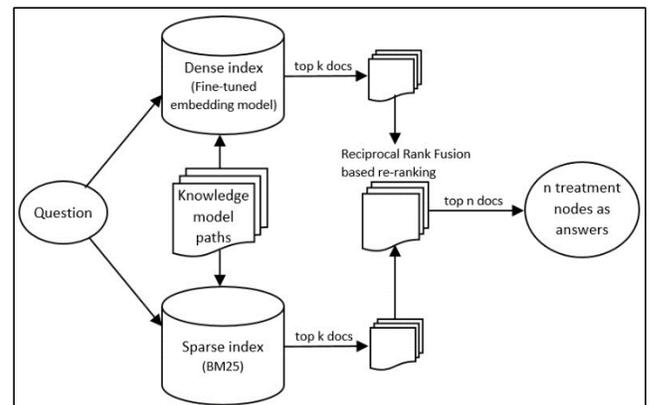

Fig. 3. Question-answering using algorithm section.

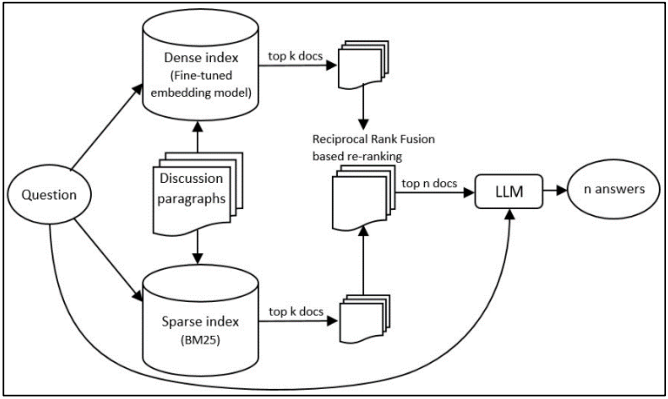

Fig. 4. Question-answering using discussion section.

### D. Fine-tuning of Embedding Model

We extracted 398 paragraphs from the treatment part of the discussion section of the NCCN guideline for NSCLC version 3.2022. A set of **1908 synthetic questions** was generated from these paragraphs by using an LLM with the appropriate prompt. The retrieval model was fine-tuned using the generated questions and the matching paragraphs to adapt the model for the domain-specific content.

### E. QA Dataset Creation

We gathered the treatment-related information from the data sources, and the potential question-answer pairs were identified manually. Except for the alterations required to convert it to Q&A format, we attempted to keep as much of the original text as possible. The question-answer pairs were saved in XML format, adapted from the MedQuAD [23] dataset. The dataset is made publicly available[1].

## III. SYSTEM IMPLEMENTATION

### A. Knowledge Extraction

We used the Apache PDFBox [14] library, version 2.0.25, to extract the graphical and textual content from the guideline documents. The textual content was extracted in HTML format by using the PDFBox library. The HTML data was then processed for paragraph-wise extraction of the text.

### B. Question Answering Framework

We used a fine-tuned version of the BAAI General Embedding (BGE) large model [15] to generate the dense vectors. The Faiss library [17] was used to create an index of the dense vectors, and the semantically similar vectors were retrieved using the L2 distance-based similarity search functionality offered by Faiss. The sparse vectors were generated using the BM25 [18] model. We used the Okapi BM25 implementation of the Rank_bm25 [20] library. The open-source large language model Llama 2 [16] was used for the synthetic question generation and the final answer generation. Specifically, we used the Llama-2-70b-chat-hf [21], a 70B Llama 2 model fine-tuned for chat completion.

## IV. RESULTS

### A. Evaluation of Knowledge Extraction Algorithm

The evaluation result of the knowledge extraction algorithm for different cancer types is shown in Table I. We manually evaluated the algorithm's output for the following NCCN guideline documents: *Breast Cancer version 4.2022*, *Prostate Cancer version 1.2023*, *Ovarian Cancer version 1.2023*, and *NSCLC version 3.2022*. The output for the NSCLC guideline document was evaluated as part of our previous work [6], and the result is included here for completeness.

TABLE I. EVALUATION RESULT: KNOWLEDGE EXTRACTION

| Cancer Type | Total Nodes | Node Formation Error Count | Node Connection Error Count | Node Formation Accuracy |
|---|---|---|---|---|
| Breast Cancer | 680 | 3 | 15 | 99.5% |
| Ovarian Cancer | 459 | 12 | 13 | 97.4% |
| Prostate Cancer | 276 | 3 | 10 | 98.9% |
| NSCLC | 914 | 7 | 6 | 99.2% |

### B. Knowledge Model Comparison

Table II shows a snippet of the knowledge model comparison result. It assigns an edit-distance-based score between 0 and 100 to measure the similarity of the two nodes' content. A score of 40 or lower was regarded as an added or removed node, and 95 or higher was considered the same content. It's a modified node otherwise.

TABLE II. KNOWLEDGE MODEL COMPARISON RESULT

| Node Id | Status | Node Content (ver 1.2023) | Node Content (ver 1.2022) | Score |
|---|---|---|---|---|
| NSCL-2/12 | Not changed | Bronchoscopy (intraoperative preferred) | Bronchoscopy (intraoperative preferred) | 100 |
| NSCL-7/4 | Modified | Multiple lesions | Multiple metastases | 69 |
| NSCL-19/10 | Added & Removed | ERBB2 (HER2) mutation positive | PD-L1 ⩾50% and negative for actionable molecular biomarkers | |

### C. Evaluation of Question Answering Framework

The question-answering system was assessed using the 22 questions compiled from the ASCO's patient resources [13]. It generates two answers for a given question from the top two documents retrieved by the hybrid retrieval mechanism. The answers were evaluated using two methods.

- **Manual evaluation by an expert**: An expert oncologist analyzed the framework's results and determined if one of the two answers to a question was correct.

---
[1]https://github.com/pralaypati/cancer_treatment_qa

- **F1 score**: F1 score calculates the average token overlapping between predicted and actual answers, as proposed in [24]. The texts are split into tokens after removing the articles, punctuations, and white spaces; then, the F1 score is computed by counting the overlapped tokens between the two. We took the highest F1 score from all of the responses for a given question and then averaged the results over all the questions.

Table III shows the evaluation result of the Q&A framework. Using the discussion section as an augmented knowledge source, one of the answers for 18 out of 22 questions was correct. On the other hand, using the algorithm section, one of the answers for 12 out of 22 questions was correct. As per our analysis, the performance is lower when using the algorithm section because the algorithm section represents the treatment options concisely, which prevents the retriever from retrieving the most important documents. Similarly, the F1 score was also low for the algorithm section. The overall F1 score was low because of the mismatch between the generic words in the predicted answer and the ground truth.

TABLE III.  EVALUATION RESULT: Q&A

| Evaluation Type | Discussion Section | Algorithm Section |
|---|---|---|
| Expert evaluation: no. of correct answers | 18 | 12 |
| Expert evaluation: accuracy | 81.8% | 54.5% |
| F1 (with fine-tuning) | 23.9% | 19.8% |
| F1 (without fine-tuning) | 19.0% | 17.7% |

V. CONCLUSION

To make cancer CPGs queryable, we proposed an enhanced version of the automated knowledge modeling algorithm to convert the NCCN guidelines for multiple cancer types. We also proposed an algorithm for comparing the knowledge models for different versions of NCCN guidelines. We created a set of question-answer pairs and a QA framework to explore the possibility of querying the model. Using the algorithm part and discussion section of the NCCN guideline, respectively, the framework can produce the answers with 54.5% and 81.8% accuracy, suggesting a significant scope of improvement for future works. This work focused on retrieving treatment options from the knowledge model for Non-Small Cell Lung Cancer; however, retrieving cancer evaluation recommendations also need to be studied. Also, the study should be broadened to encompass additional types of cancer.

ACKNOWLEDGMENT

We gratefully acknowledge Dr. Rachna Jain, Consultant, Department of Radiation Oncology, Fortis Healthcare, Shalimar Bagh, Delhi, for her invaluable assistance in evaluating the responses generated by our Q&A system.